\documentclass{article}

\usepackage{arxiv}

\usepackage[utf8]{inputenc} 
\usepackage[T1]{fontenc}    
\usepackage{hyperref}       
\usepackage{url}            
\usepackage{booktabs}       
\usepackage{amsfonts}       
\usepackage{nicefrac}       
\usepackage{microtype}      
\usepackage{lipsum}		
\usepackage{graphicx}
\usepackage{natbib}
\usepackage{doi}

\usepackage{array}
\usepackage{tabularx}
\usepackage{float}
\usepackage{changepage}
\usepackage{amsmath} 

\title{A Large-Scale AIS Dataset from Finnish Water}

\author{ \href{https://orcid.org/0009-0008-1519-2830}{\includegraphics[scale=0.06]{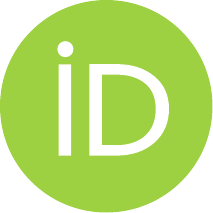}\hspace{1mm}Debayan Bhattacharya} \\
	Åbo Akademi University\\
	Erasmus Mundus Master's programme EDISS \\
	Turku, Finland \\
	\texttt{debayan.bhattacharya@abo.fi} \\
	\And
    Ikram Ul Haq \\
	Åbo Akademi University\\
	Erasmus Mundus Master's programme EDISS \\
	Turku, Finland \\
	\texttt{Ikram.UlHaq@abo.fi} \\
	\And
    Carlos Pichardo Vicencio \\
	Åbo Akademi University\\
	Erasmus Mundus Master's programme EDISS \\
	Turku, Finland \\
	\texttt{Carlos.PichardoVicencio@abo.fi}
    \And
    \href{https://orcid.org/0000-0002-5286-5343}{\includegraphics[scale=0.06]{orcid.pdf}\hspace{1mm}Sébastien Lafond} \\
	Åbo Akademi University\\
	Faculty of Science and Engineering \\
	Turku, Finland \\
	\texttt{sebastien.lafond@abo.fi} \\
}

\renewcommand{\shorttitle}{A Large-Scale AIS Dataset from Finnish Water}

\begin{document}
\maketitle

\begin{abstract}
This research paper contributes to the maritime research community by introducing a comprehensive AIS dataset from Finnish waters, specifically the Baltic Sea region. AIS data, initially designed for collision prevention, have evolved into a versatile tool with applications across diverse maritime domains. Our paper not only curates and categorises existing AIS datasets but also introduces a collected AIS dataset from the Baltic Sea area, renowned for its intercontinental cargo routes, military activities, and frozen water expanses. This dataset includes 229 millions data points and provides researchers with a resource for studying maritime activities and vessel behaviour in this dynamic region, notably distinguished by the inclusion of data from Finnish lakes. Our analysis includes a detailed listing of ship types and relevant features, empowering researchers to explore various maritime domains. To enhance comprehension and analysis, we provide visualisations of maritime traffic patterns. By publishing this AIS dataset, we aim to catalyse innovation and collaboration in maritime research, offering a gateway to deeper insights into maritime activities in the Baltic Sea and Finnish lakes.
\end{abstract}


\section{Introduction}

The maritime community has long recognised the indispensable role of Automatic Identification System (AIS) data in enhancing safety, surveillance, and navigation. To deepen this understanding, this paper provides an exploration of AIS datasets, their multifaceted applications, and presents a large AIS dataset focused on Finnish waters. This comprehensive dataset, which spans 20 months from April 2021 to December 2022, encapsulates a total of two hundred and twenty-nine million data points.

Maritime professionals, researchers, and enthusiasts have harnessed AIS data for various applications, from improving navigation safety to monitoring maritime traffic and facilitating collision avoidance. Our study highlights these various domains, emphasising the pivotal role AIS data play in optimising maritime applications. It underscores the broad relevance of AIS datasets, making them a valuable resource for anyone interested in maritime activities.

At the heart of our research is a AIS dataset designed to meet the specific needs of the maritime research community. This dataset offers an in-depth examination of vessel movements, trajectories, and contextual information. While it captures the dynamic maritime activities of the Baltic Sea, a region renowned for its intercontinental cargo exchange, military presence, and the unique challenges posed by frozen waters, it also extends its reach to the intricate network of Finnish lakes. This dataset, with its unique geographical focus, highlights the complexities of maritime activities within Finnish waters. 

We offer a glimpse of traffic patterns in our AIS dataset, showcasing the wealth of information it can reveal. Releasing this AIS dataset can advance maritime research by providing access to valuable information that supports safer, more efficient navigation and decision-making. With this dataset, we invite researchers and practitioners to explore, analyse, and innovate by harnessing the rich data resources provided.

\section{Use of AIS data in the literature}

In this section, we present an overview of the utilisation of AIS data in machine learning-based approaches as documented in the existing literature. This overview aims to equip readers with a comprehensive understanding of the achievements and advancements enabled by the use of AIS data.

AIS data has been used in various ways to conduct research across different domains. AIS data in the marine environment gives a complete view of many factors which can be further used in different application domains. AIS data can provide high-level and low-level views of a vessel's particulars in a water body.

AIS data can be used to investigate historical events and train predictive models.

Machine learning technologies can unlock new paradigms for exploiting AIS data. Furthermore, fusing AIS data with heterogeneous sensor data, including RADAR, LiDAR, and cameras, enables a comprehensive assessment of a ship's situational awareness. By applying AI and machine learning algorithms, the combined dataset can be processed to yield insightful and significant outcomes.


\subsection{Methodology}

For our literature review, we conducted a systematic search for relevant articles using the Web of Science online service. Our search centred on "AIS," and we used the keywords "navigation," "maritime," and "machine learning" across all fields. To ensure inclusiveness, we considered articles published prior to 2023 for our analysis. Consequently, we identified and selected 35 papers for our literature review. Table~\ref {tabpapers} shows the distribution of publication types obtained and analysed.

\begin{table}[] 
\caption{Table showing the distribution of document types of reviewed documents.\label{tabpapers}}
\newcolumntype{C}{>{\centering\arraybackslash}X}
\begin{tabularx}{\textwidth}{CCC}
\toprule
\textbf{Document types}	& \textbf{Record Count}	& \textbf{Percentage}\\
\midrule
Journal Article		& 25			& 75.757\%\\
Proceeding paper		& 10			& 28.571\%\\
\bottomrule
\end{tabularx}
\end{table}

\subsection{Results}
\begin{figure}[]
\includegraphics[width=13.5 cm]{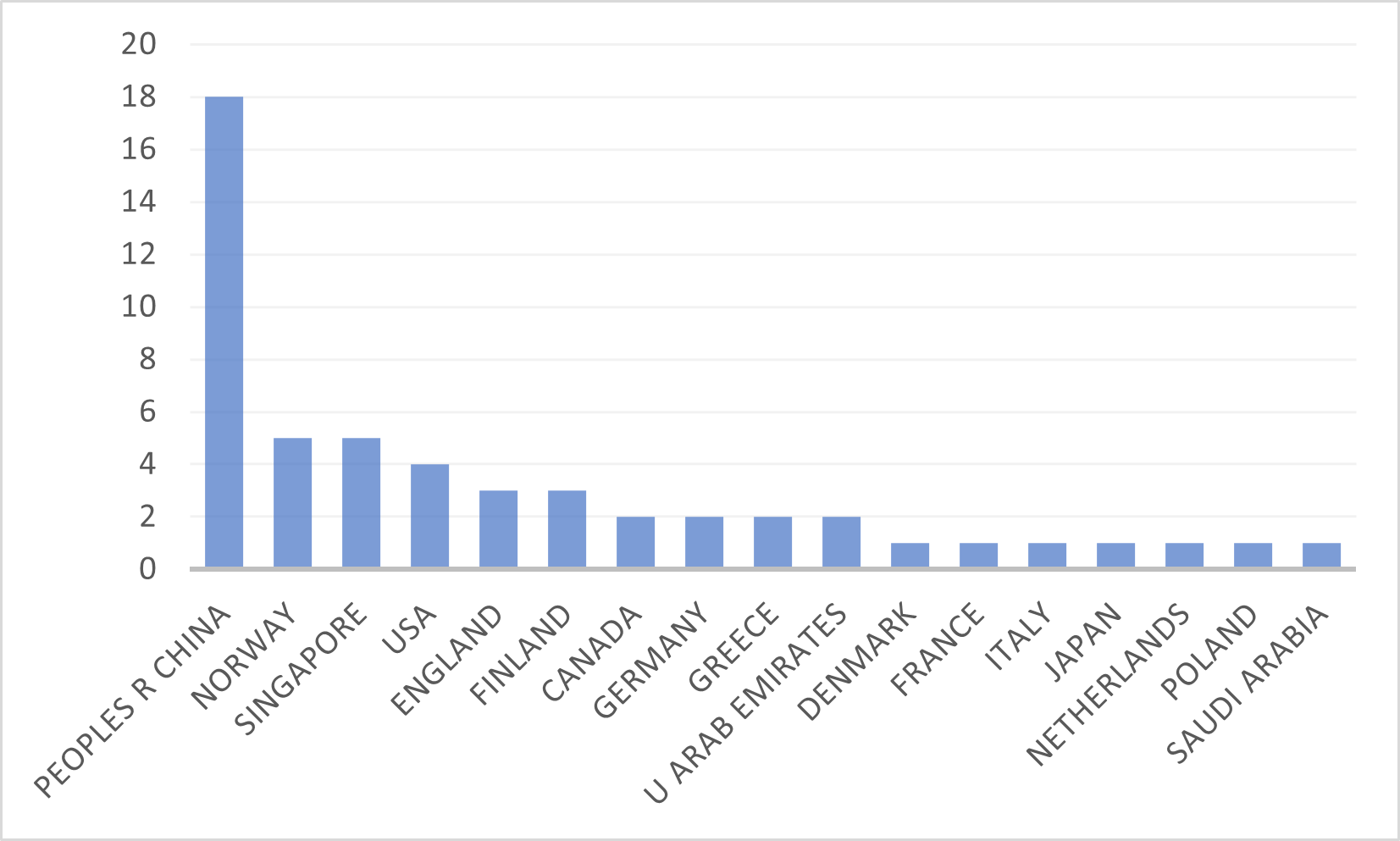}\centering
\caption{Amount of publication per country\centering \label{fig1}}
\end{figure}

In the compiled list of publications, the disciplines contributing the most papers were Marine Engineering, Oceanography, Ocean Engineering, Electrical and Electronic Engineering, and Telecommunications Engineering. Analysing the distribution of publication origins, China accounted for the largest share, with 54 per cent of publications. Norway and Singapore followed, each representing 15 per cent of the publications. Figure \ref{fig1} visually represents the publication counts corresponding to the respective countries of origin within the obtained list of publications.

\begin{figure}[]
\includegraphics[width=10.5 cm]{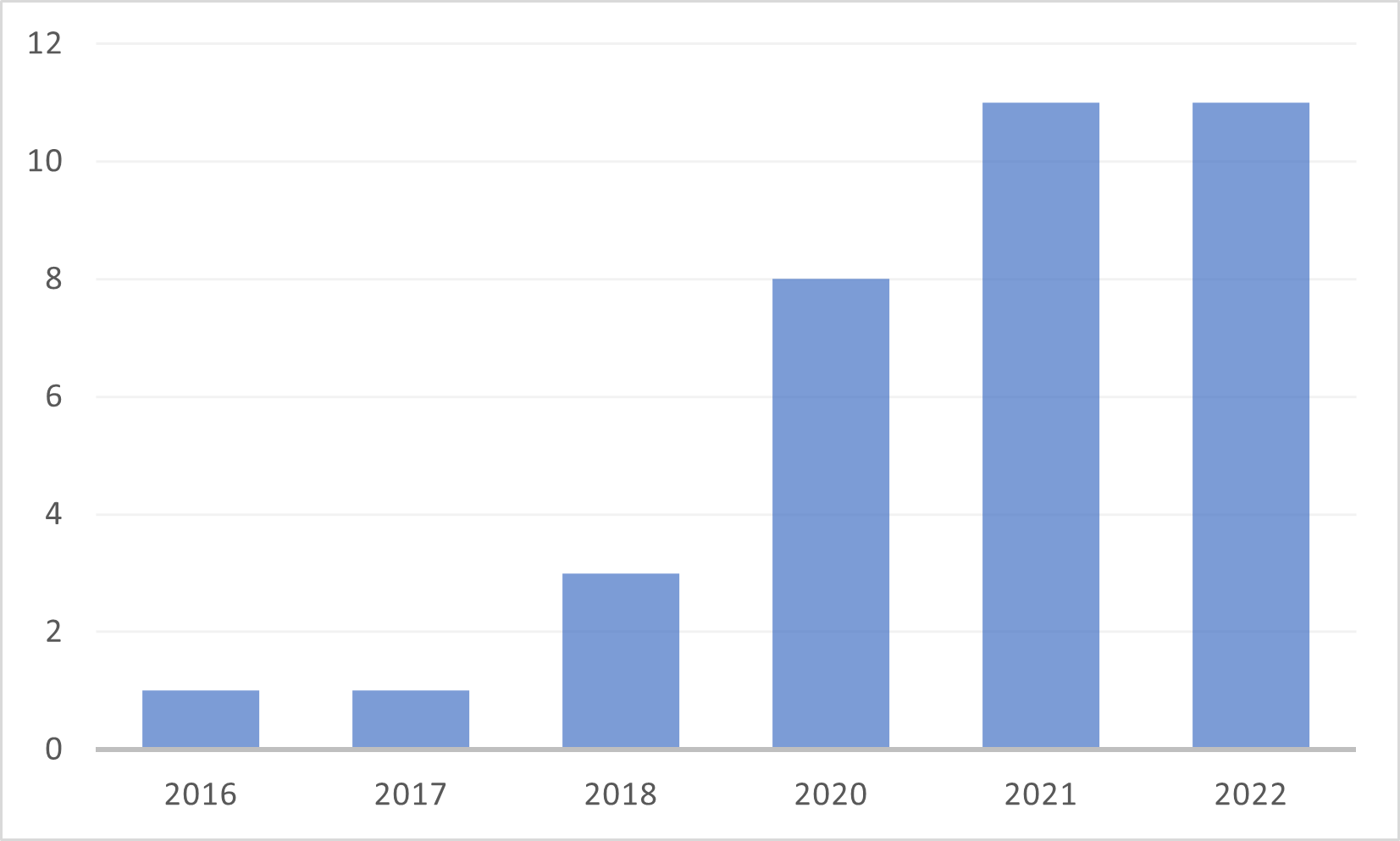}\centering
\caption{Amount of publication per year\centering \label{fig2}}
\end{figure}

Our investigation also uncovered a notable upward trend in publication volume over time. Initially, the number of publications remained minimal, with only one each in 2016 and 2017. However, this figure surged to 11 publications in 2022 alone. This progression reflects growing global research interest in the subject, making it a highly relevant and important area of study for the future. Figure \ref{fig2} displays the distribution of retrieved articles according to their respective publication years.

\subsection{Application domains}

The 33 retrieved publications span diverse maritime application domains, as outlined in Table \ref{tab3}. Most of the published works focus on advancing vessel trajectory determination methods, followed by a focus on collision-avoidance approaches. The following part of this section provides an overview of the works for each defined maritime application domain.

\begin{table}[] 
\caption{Application domains and corresponding papers.\label{tab3}}
\newcolumntype{C}{>{\centering\arraybackslash}X}
	\begin{tabularx}{\textwidth}{CCCC}
		\toprule
		\textbf{Maritime Application}	& \textbf{Publications}\\
		\midrule
		Collision Avoidance		& \cite{Shi2020} \cite{Tritsarolis2022} \cite{Chen2018}	\cite{MingyangZhang} \cite{Nishizaki2018} \cite{9187823}\cite{author2021dual} \cite{azimuth}\cite{RiskCOLREG}\cite{zhu2022}\cite{DataBasedAIS}\\
		Vessel Trajectory 		& \cite{Duan2022} \cite{Wei} \cite{Murray2022} \cite{Murray2021}  \cite{Li2020} \cite{Liu2022} \cite{Gao2018} \cite{Liu_2022} \cite{Liu__2021} \cite{Frederik} \cite{Chen2020} \cite{robot} \\
  		Ship Handling Behaviour Patterns		& \cite{sidibe_shu_2017} \cite{Gao2020}	\cite{DBSCAN2021}	\\
  		Ship grounding risk assessment		& \cite{Rawson2021} \cite{Zhang2022} \\
		Ship Sensors 		& \cite{Thombre2022} \cite{wuhan}	\\
            Port Type Prediction		& \cite{port-type}	\\
  		Risk Analysis		& \cite{Rawson2022}	\\
		Anomaly Detection		& \cite{Singh2020} \cite{Li2017}	\\
		Accident Analysis		& \cite{Liu_2021}	\\
  
	\bottomrule
	\end{tabularx}
    \end{table}

\subsubsection{Collision Avoidance}
For autonomous vessels, collision avoidance remains the primary concern \cite{Seb4}. We have already achieved a substantial level of autonomy for ground vehicles, but autonomous maritime vessels are still in earlier stages of development. This problem aims to predict vessel trajectories using historical AIS data, identify crossing points between two trajectories where vessels might collide, and take precautionary measures to avoid such situations.

Researchers have contributed to this problem using AIS datasets. A double-gated recurrent unit neural network (GRU-RNN) was constructed to learn unmanned surface vehicle (USV) collision-avoidance decisions from extracted data on successful ship encounters \cite{Shi2020}. Recently,  a team of researchers has trained a Multi-Layer Perceptron with two hidden layers having 256 and 32 neurons on default parameters provided by sklearn library \cite{Tritsarolis2022}. Nishizaki and Terayama solved the problem in two steps. First, they performed a binary classification between vessels entering the port directly and those entering the bay. Second, they predicted the direction of the exit course. They classified exit courses as turning 'right', 'left', or going straight. They used a Support Vector Machine Classifier to predict the vessel's future course \cite{Nishizaki2018}. The AIS data can be pre-processed for further use in interesting use cases. In \cite{Chen2018}, the pre-processed data is used with genetic algorithms and artificial fish swarm algorithms to find an optimal angle that allows the ship to turn safely. Previously developed human-centred methodologies like the COLREGs are now being developed with machine learning and deep reinforcement learning. The model-free development and self-learning capabilities of deep reinforcement learning make it a strong choice for autonomous vessel development \cite{RiskCOLREG}.

In recent years, researchers have utilised Automatic Identification System (AIS) data to analyse and classify various types of vessel encounters. One approach is to use ship encounter networks, which observe actions taken during collisions and use that information to inform decisions for vessels in similar situations \cite{zhu2022}.

Researchers have also developed a data-mining method in which collisions are detected by clustering ship trajectories. This research uses AIS data along with a hydro-meteorological conditions data stream. The K-means algorithm is first used to classify ship trajectories into four clusters. This is done using the static voyage features (departure, destination, voyage, and length). Then the DBSCAN algorithm is used to re-cluster using the dynamic navigation data (ship speed, course, etc.) The results are then used for collision scenario detection, followed by creating a collision risk index estimation and analysis \cite{MingyangZhang}. Researchers have also used deep learning algorithms like Proximal Policy Optimisation to enable an autonomous vessel to follow a desired path while avoiding collisions with other vessels \cite{9187823}. Researchers have also used historical AIS data with deep generative models and a dual linear autoencoder approach to predict a vessel's future trajectory. This prediction has also helped detect collisions\cite{author2021dual}. In \cite{azimuth}, a novel approach partitions ship encounter azimuth maps by integrating Automatic Identification System (AIS) data with Support Vector Classification. The core concept leverages AIS information to categorise ship encounters into distinct sectors based on their azimuth angles. This method not only yields valuable insights into the intricate web of maritime traffic dynamics but also offers substantial potential for refining collision avoidance tactics. In \cite{DataBasedAIS} AIS data features relative distance, relative speed and the course difference between two individual ships are used in a Hidden Markov Model to make an intelligent model to predict ship encounter intention.

\subsubsection{Vessel Trajectory Prediction}
Determining a vessel's trajectory has long been a major challenge. Trajectory prediction can identify travel patterns and predict future accidents and collisions. Historical AIS data can effectively support automated risk estimation, where collected points can be converted into voyages and near-real-time features to assess risk \cite{Wei}. Ship trajectories store extensive data across various time frames, such as speed and positional data. AIS data serves as a perfect dataset for leveraging machine learning methods to identify features and patterns in maritime traffic analysis \cite{Liu_2022}. With Machine learning, we can match patterns across data points through classification. Examples include K-NN classification, neural networks, and support vector machines. The AIS data contains several important parameters, both dynamic and static. All of these can improve classification accuracy \cite{Chen2020}. Recently, researchers used a semi-supervised machine learning technique to predict vessel trajectories. They used SSL-VTC, a semi-supervised deep learning approach for vessel trajectory classification based on AIS historical data \cite{Duan2022}. Murray and Perera clustered similar ship trajectories from the AIS dataset using DBSCAN. To check the similarity of a new Vessel trajectory with a cluster, a deep neural network is used to classify it among multiple clusters. After categorising the vessel trajectory, ML models can be trained for a specific cluster to predict collision points and actions to avoid them \cite{Murray2021}. Later, the same researchers divided trajectories into clusters using Gaussian mixture model clustering so that trajectories with commonalities would fall into the same cluster. They then classified a selected vessel into one of these clusters based on its observed behaviour. This way, using trajectories, precautionary measures can be taken to avoid collisions \cite{Murray2022}.

In another approach, researchers used a deep convolutional autoencoder to measure vessel trajectory similarity, and as a preprocessing step, they transformed the trajectory data into a grid \cite{Li2020}.  Some research has also focused on developing software analytical systems to handle the large data streams required for vessel trajectories for maritime situational awareness\cite{robot}. Another approach extends a spatio-temporal multigraph convolutional layer into a multigraph structure to predict vessel trajectories using a multigraph convolutional network \cite{Liu2022}.

Reconstructing a vessel's trajectory is as important as predicting it. This process allows examination and understanding of a vessel's movements over time. This can benefit researchers in areas such as maritime safety and security, environmental monitoring, economic analysis, and fleet management. 
Researchers have used vessel location, speed, and direction data to reconstruct ship trajectories. By applying a Long Short-Term Memory model and implementing a Bidirectional LSTM Recurrent Structure, the research has estimated geometric distance and reconstructed vessel paths \cite{Liu__2021} \cite{Gao2018} \cite{Frederik}. The model can help detect outliers in the acquired data, facilitating improved record-keeping for future reference.

\subsubsection{Ship handling behavior patterns}
Identifying ship-handling behaviour patterns can be beneficial in a variety of contexts. Examining typical movement patterns along various travel routes makes it possible to identify unusual patterns that may indicate vessel issues or improper handling. This can help enhance voyage security by detecting problems based on the ship's movement, as well as support environmental monitoring by identifying areas of high traffic or potential ecological impacts. Leveraging machine learning algorithms, data mining, and pattern recognition technologies, we can identify abnormal shipping behaviours to detect any illegal activities, security breaches or accidents \cite{sidibe_shu_2017}. In \cite{Gao2020} AIS trajectories are segmented to generate sub-trajectories and defined through 7-tuple coding. This approach is also used for dimensionality reduction, and data visualisation is performed using the t-distributed stochastic neighbour embedding (t-SNE) algorithm. Sub-trajectory clustering based on the spectral clustering algorithm is used for behaviour pattern recognition. The identified segments define unique ship-handling behaviour and are referred to here as ship-handling behaviour basic (SHBB).

 In another work, the authors proposed a method for classifying encounters based on AIS data; one application is using density-based spatial clustering of applications with noise (DBSCAN) to model vessel behaviours from trajectory point data. This approach clusters behaviours based on vessel profiles and can be applied to new AIS data to monitor crewed vessels and identify anomalous behaviours. An example of this type of research can be found in \cite{DBSCAN2021}.

\subsubsection{Ship grounding risk assessment}
Ship grounding risk assessment is a critical component of maritime safety and navigation. It involves evaluating the likelihood and potential consequences of ships running aground in waterways such as ports, coastal areas, and shipping lanes. This assessment is essential to mitigate the risks of groundings, which can cause environmental damage, vessel damage, and even human casualties. To conduct a thorough risk assessment, maritime experts and authorities consider factors such as navigational conditions, vessel characteristics, weather conditions, and underwater hazards. Advanced technologies, including Geographic Information Systems (GIS) and machine learning algorithms, are increasingly used to model and predict grounding risks more accurately. By identifying high-risk areas and implementing proactive measures, such as improved navigation aids and updated charts, the maritime industry aims to reduce the likelihood of ship groundings and their associated consequences, ensuring safer and more efficient maritime operations.
In \cite{Zhang2022}, the authors assessed ship grounding risk. They reconstructed trajectories from the AIS dataset using Douglas-Peucker's algorithm, then grouped them into clusters based on their behaviours. In \cite{Rawson2021} datasets are used with machine learning techniques, and this study presents a spatial maritime risk model leveraging a Discrete Global Grid System (DGGS) within this framework. The research introduces a Random Forest algorithm designed to forecast both the occurrence and spatial pattern of groundings. Impressively, the model achieves an R-squared value of 0.55 and a mean squared error of 0.002, demonstrating its robust predictive performance.

\subsubsection{Ship Sensors}
Sensors are the most fundamental units that produce valuable data for autonomous vessels \cite{Seb1,Seb2,Seb3}. Multiple sensors provide data across a wide spectrum. Sensors that give precise location data, Global Navigation Satellite System (GNSS) receivers along with Inertial Measurement Units (IMU), are often grouped with visual and audio sensors and, for remote sensing, RADAR (Radio Detection and Ranging) and LiDAR (Light Detection and Ranging)\cite{Thombre2022}. These sensors' data are also often displayed using technologies like augmented reality to improve understanding of the ship's surroundings \cite{wuhan}. These processed and collected data are the main basis for managing autonomous vessels.

\subsubsection{Port Type Prediction}
AIS data has also been used to detect and predict port type to improve vessel movement. Supervised classification algorithms like KNN (K-Nearest Neighbours) and Random Forest classifiers have been used in \cite{port-type}.

\subsubsection{Risk Analysis}
In the context of maritime risks, several machine learning algorithms are now used to predict or analyse them. In \cite{Rawson2022}, a logistic regression model is used, followed by an artificial neural network. The study also describes various datasets for addressing different maritime risk scenarios.

\subsubsection{Anomaly Detection}
Maritime anomaly detection is a vital component of maritime security and operational efficiency. It uses advanced technologies and data analysis to identify unusual or unexpected events and behaviours in the maritime domain. These anomalies can encompass a wide range of activities, from irregular vessel movements and unauthorised entries into restricted areas to deviations from established traffic patterns. Maritime anomaly detection aims to enhance safety, security, and environmental protection by quickly recognising and responding to potential threats or incidents. Leveraging data from sources such as Automatic Identification System (AIS) data, radar, and satellite imagery, this field employs sophisticated algorithms to continuously monitor maritime traffic and identify deviations from expected norms. By detecting anomalies in real time, authorities and organisations can take prompt actions to prevent accidents, illegal activities, and other undesirable events at sea, ultimately contributing to safer and more secure maritime operations. In \cite{Singh2020}, researchers used an ANN-based anomaly detection framework, and in \cite{Li2017}, they clustered similar areas of AIS point data using DBSCAN to detect anomalies in maritime activities.

\subsubsection{Accident Analysis}
Accident analysis in the maritime sector, enriched by AIS data, provides a robust framework for understanding, preventing, and mitigating maritime incidents. AIS data, including vessel positions, speeds, and course information, provides a comprehensive view of vessel movements. By leveraging this data, maritime authorities and researchers can conduct in-depth analyses of maritime accidents, such as collisions, groundings, and near-miss incidents. This detailed insight helps identify contributing factors, including human error, adverse weather conditions, or equipment malfunctions. By studying these incidents, the maritime industry can implement targeted safety measures, enhance navigational procedures, and develop more effective risk management strategies, thereby promoting safer and more efficient maritime operations. Integrating AIS data into accident analysis has become an invaluable tool for improving maritime safety and reducing the likelihood of accidents. In \cite{Liu_2021}, an advanced machine learning approach is employed to construct a data-informed Bayesian Network (BN) for the analysis of significant maritime incidents occurring in the coastal waters of China.

\subsection{Other available AIS datasets}
AIS datasets have become essential tools for studying ship movements, traffic patterns, and marine dynamics in maritime research and data analysis. This section summarises previously published AIS datasets, covering both free and paid options, and provides researchers with information about prior work.

Table \ref{tab3}  lists information on available AIS datasets retrieved by the authors through the use of AIS data in the literature.
These datasets offer a broad view of ship positions, identifying details, and related features, making them crucial resources for research ranging from transportation optimisation to marine domain awareness. As they explore this area, researchers and analysts frequently look for freely accessible AIS information to support their findings.

\begin{table}[] 
\caption{Available AIS datasets.\label{tab3}}
\newcolumntype{C}{>{\centering\arraybackslash}X}
	\begin{tabularx}{\textwidth}{CCCC}
		\toprule
		\textbf{Dataset}	& \textbf{Feely Available}\ & \textbf{Paid}	& \textbf{URL}\\
		\midrule

        US Coast Guard Navigation Center & Yes & No & \href{https://marinecadastre.gov/ais/}{Link} \\
        Fleet Mon & No & Yes & \href{https://www.fleetmon.com/services/ais-data-shop/passenger/}{Link} \\
        Marine Traffic & on Request & No & \href{https://www.marinetraffic.com/research/open-research-maritime-vessel-tracking-dataset/}{Link} \\
        Vessel Tracker & No & Yes & \href{https://www.vesseltracker.com/en/products/dataServices.html}{Link} \\
        Historical AIS messages from vessels in Danish water & Yes & No & \href{https://dma.dk/safety-at-sea/navigational-information/ais-data}{Link} \\
        The 2015 Tianjin Port AIS dataset & No & N/A & N/A \\
		\bottomrule
	\end{tabularx}
\end{table}



\section{Proposed AIS dataset}

The proposed AIS dataset covers a substantial 20-month period, from April 2021 to December 2022. This extensive coverage period empowers analysts to examine long-term trends and variations in vessel activities. Moreover, it helps researchers understand the potential influence of external factors, including weather patterns, seasonal variations, and economic conditions, on vessel traffic and behaviour in Finnish waters.
This dataset includes an extensive array of data on vessel movements and activities across seas, rivers, and lakes. Anticipated to be comprehensive, the dataset includes a diverse range of ship types, such as cargo ships, tankers, fishing vessels, passenger ships, and other categories.

The AIS dataset features exceptional granularity, with a total of 2 293 129 345 data points. Providing such granular information can help analysts understand vessel dynamics and operations in Finnish waters. It enables identification of patterns and anomalies in vessel behaviour and supports assessment of the potential environmental implications of maritime activities.

We obtained the dataset via the public MQTT WebSocket APIs offered by Fintraffic traffic services. The data we present includes two CSV files per day: the location file (location.csv) and the metadata file (metadata.csv). The dataset is freely available from {\cite{zenodo} }.

\subsection{Details about location file}
Each location file includes a wide range of essential features that provide comprehensive insights into vessel positions in Finnish waters. These features include timestamps, timestampExternal, MMSI numbers, longitude (lon), latitude (lat), speed over ground (sog), course over ground (cog), navigational status (navStat), rate of turn (rot), position accuracy (posAcc), RAIM flag (raim), and true heading. This file contains the received AIS position reports. The structure of the logged parameters is the following: [timestamp, timestampExternal, mmsi, lon, lat, sog, cog, navStat, rot, posAcc, raim, heading]

\textbf{Timestamps:} the timestamp feature in our location .csv file represents the Universal Coordinated Time (UTC) second when the report was generated by the electronic position system (EPFS). It ranges from 0 to 59, indicating the precise moment at which the report was generated. In cases where the timestamp is not available, which is considered the default value, it is represented as 60. Additionally, a value of 61 signifies that the positioning system is in manual input mode, while a value of 62 indicates that the electronic position fixing system operates in estimated (dead reckoning) mode. Finally, if the positioning system is inoperative, the timestamp is recorded as 63. These distinct values provide essential information regarding the operational status and accuracy of the electronic position system used to generate the report.

\textbf{TimestampExternal:} the parameter represents the timestamp associated with the MQTT message received from www.digitraffic.fi. It is assumed that this timestamp corresponds to the Epoch time when the AIS message was received by digitraffic.fi

\textbf{MMSI:} the parameter refers to the Maritime Mobile Service Identity number assigned to a vessel. It is a unique 9-digit identifier used in the Digital Selective Calling (DSC) radio or AIS unit for vessel identification and communication.

\textbf{Longitude (lon):} the longitude parameter provides the longitude of a vessel's position. It is represented in 1/10,000 minutes and ranges from -180 degrees (west) to +180 degrees (east). A value of 181 (6791AC0h) indicates that the longitude information is not available or is set as the default.

\textbf{Latitude (lat):} the latitude parameter represents the latitude of a vessel's position. It is also expressed in 1/10,000 minutes and ranges from -90 degrees (south) to +90 degrees (north). A value of 91 degrees (3412140h) indicates that the latitude information is not available or is set as the default.

\textbf{Speed over Ground (SOG):} the parameter denotes the vessel's speed relative to the Earth's surface. It is measured in 1/10 knot steps and ranges from 0 to 102.2 knots. A value of 1023 indicates that the speed over ground is not available, while a value of 1022 or higher indicates a speed of 102.2 knots or more.

\textbf{Course over Ground (COG):} the parameter represents the direction of a vessel's movement over the ground. It is measured in 1/10 degrees and ranges from 0 to 3599. A value of 3600 indicates that the course over ground is not available or is set as the default.

\textbf{Navigational Status (navStat)} the parameter provides information about the vessel's current operational status. It is represented by numeric codes as follows: 0 (under way using engine), 1 (at anchor), 2 (not under command), 3 (restricted maneuverability), 4 (constrained by her draught), 5 (moored), 6 (aground), 7 (engaged in fishing), 8 (under way sailing), 9 (reserved for future amendment of navigational status), 10 (reserved for future amendment of navigational status), 11 (power-driven vessel towing astern), 12 (power-driven vessel pushing ahead or towing alongside), 13 (reserved for future use), 14 (AIS-SART, MOB-AIS, EPIRB-AIS), and 15 (undefined or default).

\textbf{Rate of Turn (ROT):} the parameter represents the vessel's rate of turn in degrees per minute. Positive values indicate turning right, while negative values indicate turning left. The range is from 0 to 126 degrees per minute, with higher values indicating turning at rates beyond 708 degrees per minute. A value of +127 indicates turning right at more than 5 degrees per 30 seconds, while -127 indicates turning left at more than 5 degrees per 30 seconds. A value of -128 indicates that no turn information is available or is set as the default.

\textbf{Position Accuracy (posAcc):} the parameter indicates the accuracy of the vessel's reported position. It is represented by a binary value: 1 for high accuracy (<= 10 meters) and 0 for low accuracy (> 10 meters). A value of 0 is also set as the default.

\textbf{RAIM Flag (raim):} the RAIM (Receiver Autonomous Integrity Monitoring) Flag parameter represents the status of the receiver's autonomous integrity monitoring for the electronic position fixing device. It is represented by a binary value: 0 for RAIM not in use (default) and 1 for RAIM in use.

\textbf{Heading:} the parameter indicates the true heading of the vessel in degrees, ranging from 0 to 359. A value of 511 indicates that the heading information is not available or is set as the default.

\subsection{Details about metadata file}

The metadata file for our AIS dataset provides essential information that enhances context and understanding of the vessel data. This file includes parameters such as timestamp, destination, MMSI, callSign, IMO, shipType, draught, ETA, posType, pointA, pointB, pointC, pointD, and name. The structure of the features are [timestamp, destination, mmsi, callSign, imo, shipType, draught, eta, posType, pointA, pointB, pointC, pointD, name] 

\textbf{Timestamp:} the parameter represents the UTC second when the report was generated by the electronic position system (EPFS). It provides a crucial chronological reference for each AIS position report. The timestamp values range from 0 to 59, indicating the exact second of the report generation. If the timestamp is not available, the default value of 60 is assigned. Additionally, certain special values are used to indicate specific conditions: 61 represents the positioning system in manual input mode, 62 indicates the electronic position fixing system operating in estimated (dead reckoning) mode, and 63 signifies an inoperative positioning system.

\textbf{Destination:} the field contains information about the intended destination of the vessel. It is represented using a maximum of 20 characters. The value "@@@@@@@@@@@@@@@@@@@@" is used to indicate that the destination is not available. In the context of SAR aircraft, the use of the destination field may be determined by the responsible administration.

\textbf{MMSI:} the parameter corresponds to the Maritime Mobile Service Identity (MMSI) number. MMSI is a unique 9-digit identifier assigned to Digital Selective Calling (DSC) radios or AIS units. It serves as a key identifier for vessels and enables communication and identification in maritime systems. The MMSI number is a crucial piece of information for vessel tracking, communication, and coordination.

\textbf{Call Sign:} the field represents the call sign of the vessel. It is a 7-character code encoded in 6-bit ASCII. The value "@@@@@@@" is used to indicate that the call sign is not available or is the default value. In the case of craft associated with a parent vessel, the call sign should use "A" followed by the last 6 digits of the MMSI of the parent vessel.

\textbf{IMO Number:} the parameter refers to the International Maritime Organization (IMO) number. It is a unique identifier assigned to seagoing ships and registered ship owners. The value "0" is used to indicate that the IMO number is not available or is the default value. The IMO number provides a globally recognized identification system for ships.

\textbf{Ship Type:} the field indicates the type of vessel. It is represented by a numeric code ranging from 0 to 199. Values from 1 to 99 are defined for specific ship types, while values from 100 to 199 are reserved for regional use. Values from 200 to 255 are reserved for future use and are not applicable to SAR aircraft. The ship type information provides insights into the characteristics and capabilities of the vessel.

\textbf{Draught:} the parameter represents the draught of the vessel in 1/10 meters. It provides information about the depth of the vessel's keel below the waterline. The value "255" is used to indicate a draught of 25.5 meters or greater, while "0" represents that the draught information is not available or is the default value. The draught information is important for vessel navigation, especially in shallow waters.

\textbf{ETA:} the field stands for the estimated time of arrival. It represents the expected time of the vessel's arrival at a specific location in UTC format (MMDDHHMM). The ETA is divided into four sections: month, day, hour, and minute. Each section has its own format, with "0" indicating that the corresponding information is not available or is the default value. The ETA provides valuable information for planning and scheduling vessel movements.

\textbf{Position Type:} the \textit{posType} parameter indicates the type of electronic position fixing device used by the vessel. It is represented by a numeric code. Some of the commonly used codes include 0 for undefined (default), 1 for GPS, 2 for GLONASS, 3 for combined GPS/GLONASS, 4 for Loran-C, 5 for Chayka, 6 for integrated navigation system, 7 for surveyed, 8 for Galileo, and 15 for internal GNSS. Values 9 to 14 are not used. The posType field provides information about the positioning technology employed by the vessel.

\textbf{Reference Points:} the \textit{pointA}, \textit{pointB}, \textit{pointC}, and \textit{pointD} fields serve as reference points for the reported position. These points also indicate the dimensions of the ship in meters. The use of these fields for SAR aircraft may be determined by the responsible administration. If utilized, the reference points should indicate the maximum dimensions of the craft. As a default, the values of pointA, pointB, pointC, and pointD are set to "0". The reference points provide additional information about the size and shape of the vessel.

\textbf{Name:} the field represents the name of the vessel. It can have a maximum of 20 characters encoded in 6-bit ASCII. The value "@@@@@@@@@@@@@@@@@@@@" is used to indicate that the name is not available or is the default value. The name should be consistent with the station radio license. For SAR aircraft, the name is set to "SAR AIRCRAFT NNNNNNN," where NNNNNNN represents the aircraft registration number. The vessel's name is an important identifier for communication and identification purposes.

\section{Overview of the AIS dataset content}
The proposed AIS dataset encompasses various ship types, each represented by a numeric code in the shipType field. These codes indicate the general category and characteristics of the vessels. The defined ship types range from 1 to 99 and include cargo ships, tankers, passenger vessels, fishing boats, and more. The ship type information enables the identification and classification of vessels by their primary purpose and operations. Understanding the different ship types in the AIS dataset helps analyse maritime traffic patterns and gain insights into the diverse activities taking place on the waterways.

We show a representation of the different ship types in our dataset, with the unique MMSI and IMO numbers in Table \ref{tabMMSI} and Table \ref{tabIMO}, respectively. In the AIS dataset, MMSI (Maritime Mobile Service Identity) numbers play a crucial role in identifying and distinguishing individual vessels. MMSI numbers are unique 9-digit identifiers assigned to maritime radio systems, including AIS units and DSC (Digital Selective Calling) radios. These numbers remain constant throughout a vessel's lifespan, providing a consistent identifier for tracking and communication purposes. However, it's important to note that MMSI numbers can change in certain circumstances, such as when a vessel changes ownership or its flag state registration.


On the other hand, IMO (International Maritime Organization) numbers serve as permanent identifiers for ships. Unlike MMSI numbers, IMO numbers are assigned to individual vessels rather than communication devices. These numbers consist of a unique seven-digit code that remains unchanged throughout a vessel's lifetime. IMO numbers are assigned to ships larger than 100 gross tons or those engaged in international voyages. They serve as a consistent identification method for vessel tracking, safety, and regulatory purposes. Unlike MMSI numbers, IMO numbers do not change with ownership or registration changes, providing a stable, reliable identifier for vessels throughout their operational lifespan.

When comparing MMSI numbers and IMO numbers, it is important to note their distinct purposes and characteristics. MMSI numbers primarily serve as communication identifiers, allowing for direct contact and message exchange between vessels and shore stations. In contrast, IMO numbers are permanent, unique identifiers assigned to individual vessels, supporting global recognition, tracking, and regulatory compliance. While both MMSI and IMO numbers support vessel identification and communication, they serve different functions and follow separate assignment and maintenance procedures in the maritime domain.

\begin{table}[] 
\caption{Distribution of ship types in the dataset based on unique MMSI numbers.\label{tabMMSI}}
\newcolumntype{C}{>{\centering\arraybackslash}X}
\begin{tabularx}{\textwidth}{CCC}
\toprule
\textbf{Ship Types}	& \textbf{No of Ships} & \textbf{Percentage}\\
\midrule
Cargo	&4170 &51.53\% \\
Tanker	&1804 &22.29\%\\
Passenger	&434 &5.36\%\\
Tug	&304 &3.75\%\\
Other Type	&266 &3.28\%\\
Not available (default)	&229 &2.83\%\\
Search and Rescue vessel	&167 &2.06\%\\
Military ops	&165 &2.039\%\\
Pilot Vessel	&144 &1.77\%\\
Dredging or underwater ops	&70 &0.86\%\\
Pleasure Craft	&70 &0.86\%\\
Law Enforcement	&67 &0.82\%\\
Sailing	&61 &0.75\%\\
Towing	&28 &0.34\%\\
Reserved for future use	&24 &0.29\%\\
Anti-pollution equipment	&19 &0.23\%\\
High speed craft (HSC)	&18 &0.22\%\\
Diving ops	&15 &0.18\%\\
Reserved	&8 &0.09\%\\
Port Tender	&8 &0.09\%\\
Wing in ground (WIG)	&6 &0.074\%\\
Noncombatant ship according to RR Resolution No. 18	&5 &0.06\%\\
Spare - Local Vessel	&4 &0.04\%\\
Medical Transport	&3 &0.03\%\\
Towing: length exceeds 200m or breadth exceeds 25m	&2 &0.02\%\\

\bottomrule
\end{tabularx}
\end{table}


\begin{table}[] 
\caption{Distribution of ship types in the dataset based on unique IMO numbers.\label{tabIMO}}
\newcolumntype{C}{>{\centering\arraybackslash}X}
\begin{tabularx}{\textwidth}{CCC}
\toprule
\textbf{Ship Types}	& \textbf{No of Ships} & \textbf{Percentage}\\
\midrule
Cargo & 3961 &59.68\% \\
  Tanker & 1731 &26.08\% \\
  Passenger & 261 &3.93\% \\
  Tug & 242 &3.64\% \\
  Other Type & 137 &2.06\% \\
  Not available (default) & 102 &1.53\% \\
  Dredging or underwater ops & 41 &0.61\% \\
  Pleasure Craft & 25 &0.37\%\\
  Towing & 22 &0.33\% \\
  Sailing & 20 &0.30\% \\
  Military ops & 15 &0.22\% \\
  Reserved for future use & 15 &0.22\% \\
  Law Enforcement & 12 &0.18\% \\
  Pilot Vessel & 11 & 0.16\% \\
  Anti-pollution equipment & 9 & 0.13\% \\
  Search and Rescue vessel & 7 &0.10\% \\
  Reserved & 6 &0.09\% \\
  High speed craft (HSC) & 5 &0.07\% \\
  Wing in ground (WIG) & 4 &0.06\% \\
  Diving ops & 3 &0.045\% \\
  Noncombatant ship according to RR Resolution No. 18 & 3 & 0.04\% \\
  Spare - Local Vessel & 2 &0.03\% \\
  Port Tender & 1  &0.01\% \\
  Towing: length exceeds 200m or breadth exceeds 25m & 1 &0.01\% \\

\bottomrule
\end{tabularx}
\end{table}

\subsection{Ship features inside Finnish Lakes}
A diverse fleet of ships and vessels plies the Finnish lakes, serving different needs and purposes. Passenger ferries carry locals and tourists, connecting lakeside communities and supporting local travel. Cargo ships keep goods moving across the lakes, supporting trade and commerce. Research vessels contribute to scientific studies and environmental monitoring, while rescue boats ensure safety and security on the water.

Several cities and towns are nestled near Finnish lakes. Helsinki, the capital city, sits on the shores of the Gulf of Finland, with easy maritime access to the archipelago. Tampere, Jyväskylä, and Kuopio are notable cities near large Finnish lakes, offering regular maritime inland activities.

Our AIS data set holds a distinctiveness that sets it apart from others, as it provides valuable insights into maritime activities within Finnish lakes. The inclusion of AIS data from these lakes offers a unique perspective on vessel movements, navigational patterns, and the overall dynamics of maritime operations in this region.

\begin{figure}[]
\includegraphics[width=13.5 cm]{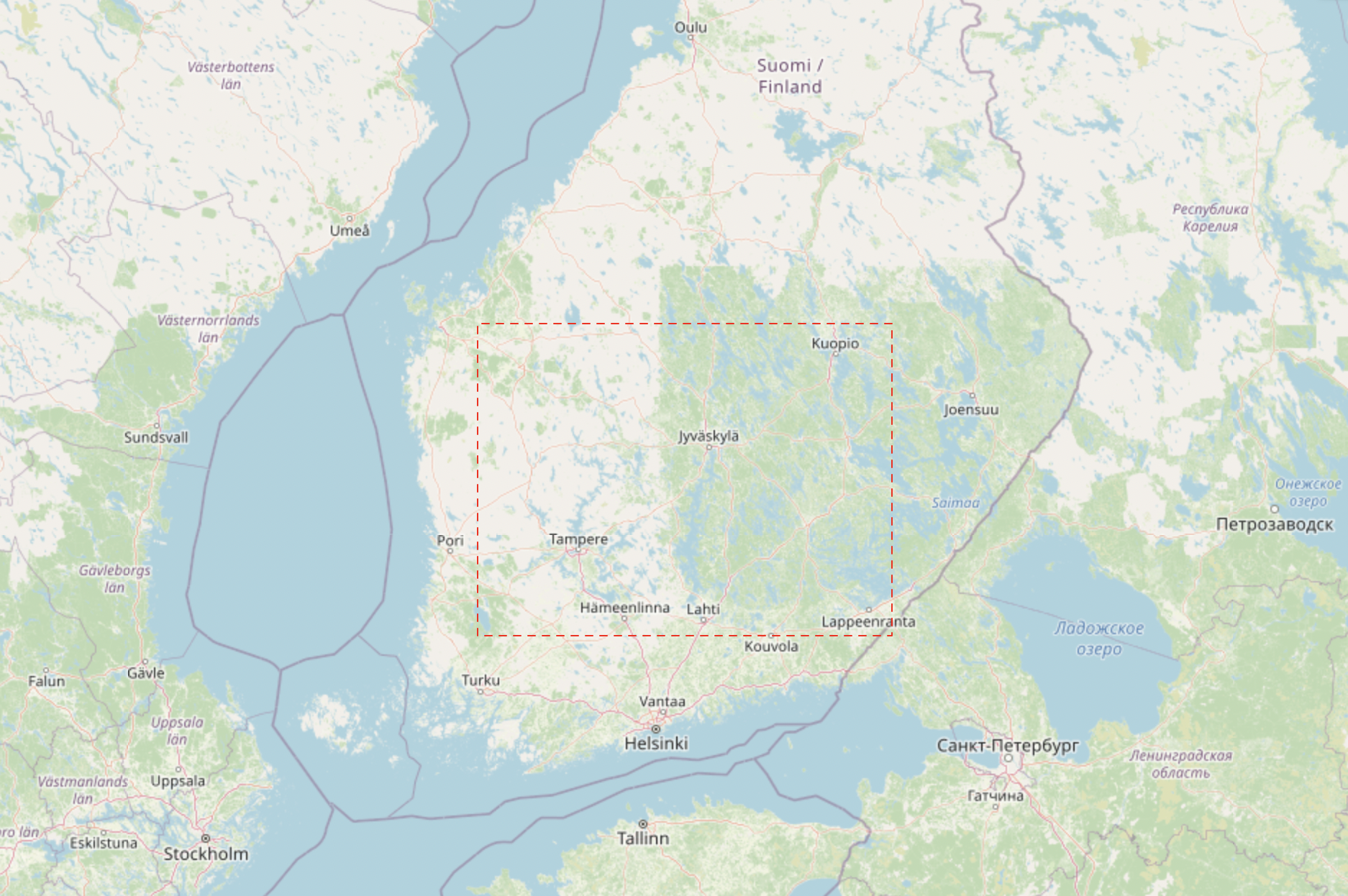}\centering
\caption{The area which we had defined to cover some of the lakes in Finland. The territory was bound with the coordinates 60.867639, 22.218472 and 63.100972, 28.549167 \centering \label{fig3}}
\end{figure}

For illustration, we used data from the dataset for June, July, and August 2021 and 2022. We defined an area within Finland's territory that encompasses lakes and rivers. The territory was bounded by the coordinates 60.867639, 22.218472 and 63.100972, 28.549167.  Figure \ref{fig3} shows the bounding box that covers some of the Finnish lakes we covered, and Table \ref{fig3} provides the distribution of ship types encountered in this defined territory.

\begin{table}[] 
\caption{Distribution of ship types in the various Finnish lakes in the dataset based on unique MMSI numbers.\label{tablakes}}
\newcolumntype{C}{>{\centering\arraybackslash}X}
\begin{tabularx}{\textwidth}{CCC}
\toprule
\textbf{Ship Types}	& \textbf{No of Ships} & \textbf{Percentage}\\
\midrule
Cargo & 66 &51.96\% \\
  Tug & 16 &12.59\% \\
  Passenger & 15 &11.81\% \\
  Search and Rescue vessel & 8 &6.29\% \\
  Other Type & 7 &5.51\% \\
  Pilot Vessel & 4 &3.14\% \\
  Towing & 2 &1.57\% \\
  Anti-pollution equipment & 2 &1.57\% \\
  Tanker & 2 &1.57\% \\
  Pleasure Craft & 1 &0.78\% \\
  Spare - Local Vessel & 1 &0.78\% \\
  Not available (default) & 1 &0.78\% \\
  Noncombatant ship according to RR Resolution N... & 1 &0.78\% \\
  136 & 1 &0.78\% \\

\bottomrule
\end{tabularx}
\end{table}

\subsection{Visualisation of the AIS dataset content}
To gain insights into the dataset, we generated heatmaps that visualise the accumulation of ship activity based on AIS data. These heatmaps reveal areas with higher concentrations of ship movements, helping us identify prominent routes connecting ports and countries. Additionally, we can analyse activity patterns in Finland's lakes and other regions, shedding light on diverse maritime endeavours in those areas.

\begin{figure}[]
\includegraphics[width=10.5 cm]{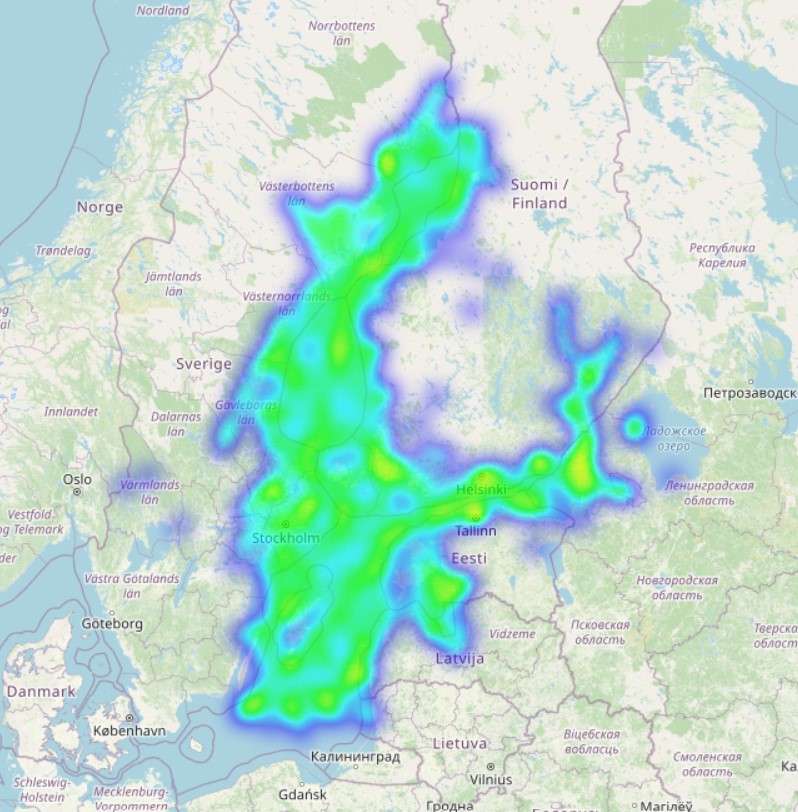}\centering
\caption{ Heatmap of the AIS data during the days of winter from December 2021 to February 2022  \centering \label{map_winter}}
\end{figure} 

This experiment utilised the Folium map library to consolidate information over a specific time period. We implemented a grid-search aggregation algorithm that reduced the number of ship position data points, allowing efficient aggregation in Folium. We decided to collect data points monthly and seasonally for the designated periods.

\begin{figure}[]
\includegraphics[width=10.5 cm]{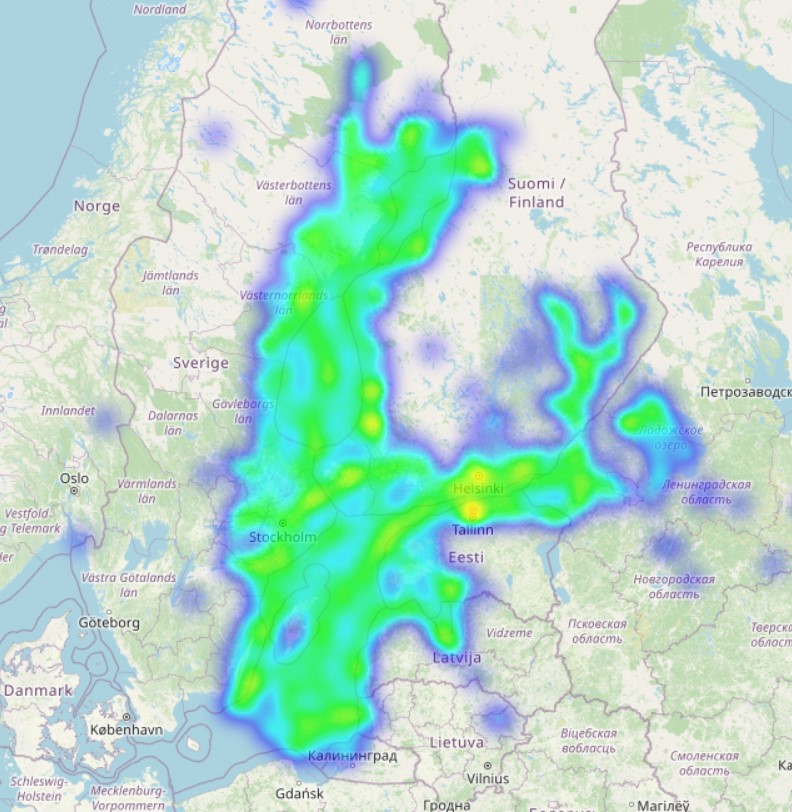}\centering
\caption{ Heatmap of the AIS data during the days of spring from March 2022 to May 2022  \centering \label{map_spring}}
\end{figure} 

\begin{figure}[]
\includegraphics[width=10.5 cm]{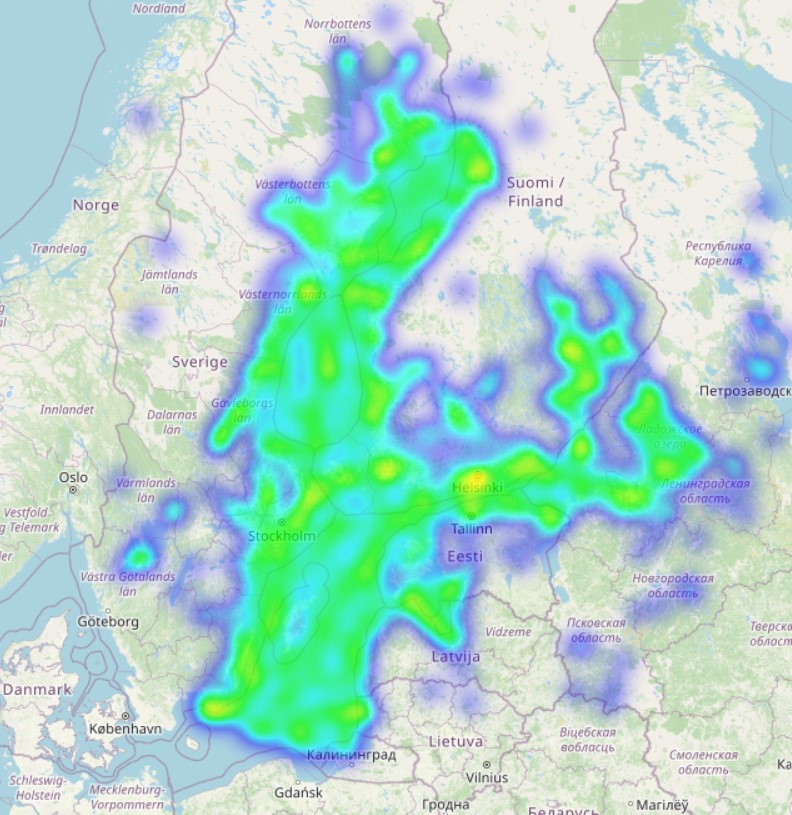}\centering
\caption{ Heatmap of the AIS data during the days of summer from June 2022 to August 2022  \centering \label{map_summer}}
\end{figure} 

\begin{figure}[]
\includegraphics[width=10.5 cm]{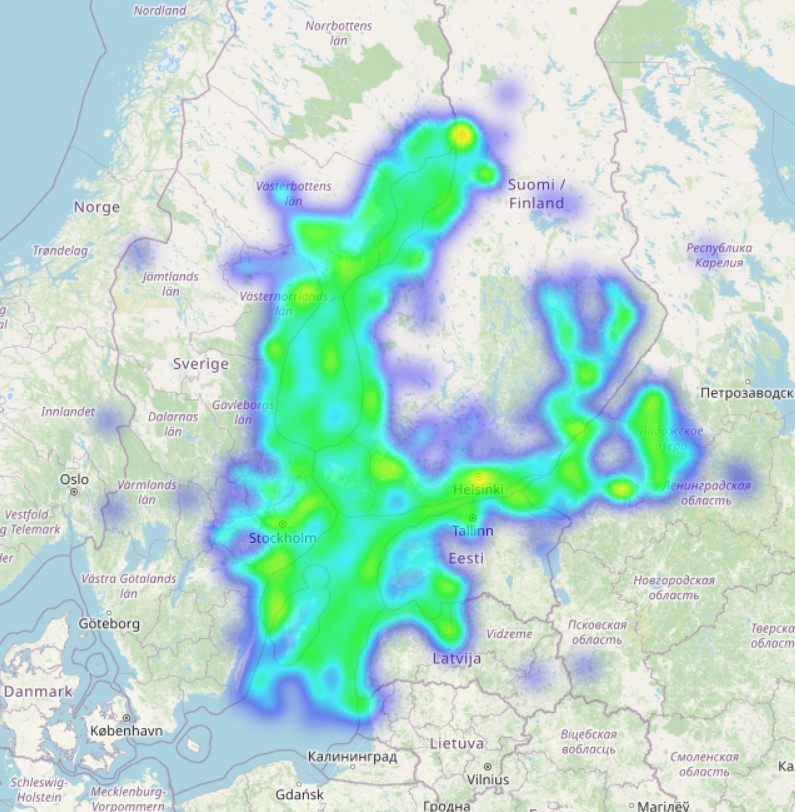}\centering
\caption{ Heatmap of the AIS data during the days of autumn from September 2022 to November 2022  \centering \label{map_autumn}}
\end{figure} 

When observing the four seasons in the heatmaps, we can see differences in activity throughout the year. First, in the winter heatmap in Figure \ref{map_winter}, we see that most activity occurs around the Baltic Sea; this activity is likely ferries and cruises across different countries, with predefined routes. As shown, activity on the lakes is rarely observed during these months, as the lakes are frozen, making it impossible to use small vessels.

In the spring heat map in Figure \ref{map_spring}, we see port activity begin to increase, with the most activity in ports like Helsinki and Tallinn, as ferry connections may start to increase. This activity is starting to spread from the main Baltic Sea to other nearby regions, with entries into major ports or near the shore.

In the summer heatmap in Figure \ref{map_summer}, you can see a large difference compared to the other maps: activity increases sharply around the different bodies of water. In the lakes, we see much more activity because of the season; small vessels like personal sailboats increase activity in these regions, with prominent activity across the map. Similarly, we can see activity spread and reach a high point, with more positions across different regions of the Baltic Sea.

The last map we can observe is the autumn heatmap in Figure \ref{map_autumn}. In this one, we can observe that, compared to the summer map in Figure \ref{map_summer}, the activity is starting to decrease and is becoming more confined to the areas of the Baltic Sea, becoming more similar to the map of spring in Figure \ref{map_spring}


Also, the maps show some points that appear outside bodies of water. These errors likely come from human error when setting up an AIS transmitter or configuring the associated GNSS receiver. These reflect real-world AIS errors and illustrate the positional inaccuracies that can occur in AIS data.


\section{Conclusion}
In conclusion, our work aims to support and strengthen the maritime research community by introducing a rich, extensive AIS dataset specific to Finnish waters in the Baltic Sea and surrounding lakes. As maritime activities continue to evolve and diversify, the need for high-quality, region-specific datasets becomes increasingly critical. By providing this dataset, we aim to empower researchers, scientists, and maritime experts to delve into the intricacies of this dynamic region. We believe our dataset can foster a deeper understanding of vessel behaviour, traffic patterns, and accident risk assessment. We hope this contribution inspires innovative research and collaboration in the maritime domain, offering insights that support safer, more efficient maritime operations. As we take this step forward, we encourage the maritime research community to explore, expand, and utilise this dataset to unlock new horizons in maritime research and safety.

\section*{Funding}{The work has been partially supported by the EMJMD master's programme in Engineering of Data-Intensive Intelligent Software Systems (EDISS - European Union’s Education, Audiovisual and Culture Executive Agency grant number 619819).}

\bibliographystyle{unsrt}

\bibliography{references}  






\end{document}